\documentclass[10pt,twocolumn,letterpaper]{article}

\usepackage{iccv}
\usepackage{times}
\usepackage{epsfig}
\usepackage{graphicx}
\usepackage{amsmath}
\usepackage{amssymb}

\usepackage{booktabs}
\usepackage{color}
\usepackage{bm}
\usepackage{makecell}
\usepackage{arydshln}
\usepackage{subcaption}
\usepackage{enumitem}

\usepackage[pagebackref=true,breaklinks=true,letterpaper=true,colorlinks,bookmarks=false]{hyperref}

\iccvfinalcopy % *** Uncomment this line for the final submission
\usepackage{caption}
\usepackage[capitalize]{cleveref}
\crefname{section}{Sec.}{Secs.}
\Crefname{section}{Section}{Sections}
\Crefname{table}{Table}{Tables}
\crefname{table}{Tab.}{Tabs.}

\def\iccvPaperID{2403} % *** Enter the ICCV Paper ID here

\ificcvfinal\fi

\begin{document}

%%%%%%%%% TITLE
\title{DriveAdapter: Breaking the Coupling Barrier of \\
Perception and Planning in End-to-End Autonomous Driving}

\author{
Xiaosong Jia$^{1,2}$,
Yulu Gao$^{2,3}$,
Li Chen$^{2}$,
Junchi Yan$^{1,2^\dagger}$,
Patrick Langechuan  Liu$^{4}$,
Hongyang Li$^{2,1^\dagger}$ \\
[2mm]
$^1$~Shanghai Jiao Tong University \quad 
$^2$~OpenDriveLab, Shanghai AI Lab \\ 
$^3$~Beihang University \quad 
$^4$~Anker Innovations 
\\
\normalsize{
$^\dagger$Correspondence authors}\\
\normalsize{
\url{https://github.com/OpenDriveLab/DriveAdapter}
}}

% \\
% \normalsize{
% \url{XXX}
% }

%\maketitle
\ificcvfinal\thispagestyle{empty}\fi

\twocolumn[{%
\renewcommand\twocolumn[1][]{#1}%
\maketitle
\begin{center}
    \centering
    \captionsetup{type=figure}
    \includegraphics[width=0.95\textwidth]{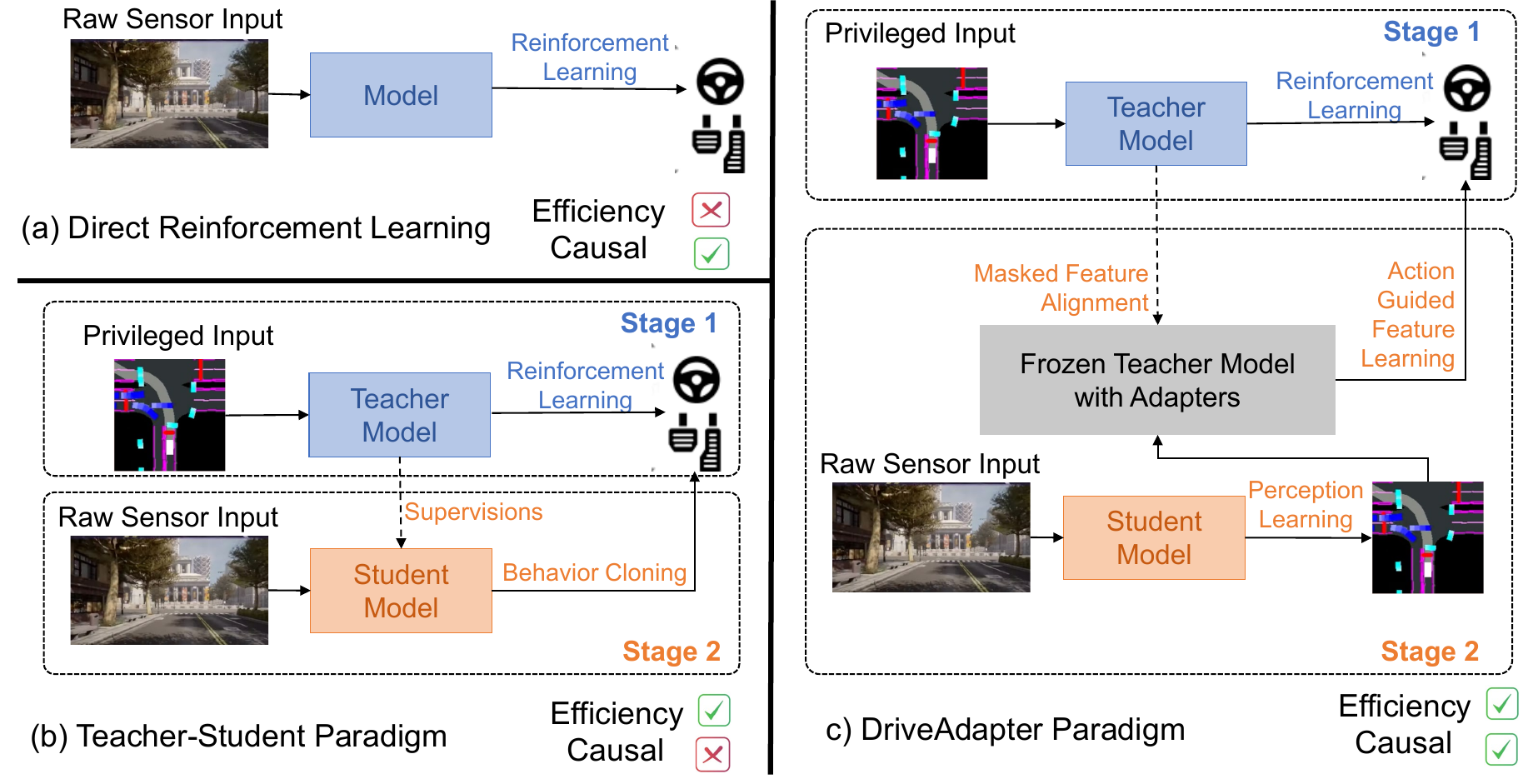}
    \vspace{-5pt}
    \captionof{figure}{\textbf{Comparison of different paradigms for end-to-end autonomous driving.} (a) Though directly conducting reinforcement learning (RL) with raw sensor inputs~\cite{toromanoff2020end} enables the inference model to learn the causal relationship of driving through rewards, it requires tens of days of training with the simulator rendering raw sensor inputs, which is of low efficiency.  (b) State-of-the-art works~\cite{chen2020learning,chen2022lav,wu2022trajectoryguided,Renz2022CORL,hu2022model} usually adopt the teacher-student paradigm to enable efficient policy learning by providing privileged inputs (perception ground-truth) to the RL model. However, their student model suffers from the causal confusion issues~\cite{wen2020fighting} due to behavior cloning. (c) In the proposed DriveAdapter paradigm, the model still enjoys high RL training efficiency while the usage of the frozen teacher model empowers the inference process with its driving knowledge. The student model could focus on perception learning, and the proposed adapter module with its masked feature alignment objective functions deals with the distribution gap between the perception results and the privileged inputs.  \label{fig:teaser}}
\end{center}}]

%%%%%%%%% ABSTRACT
\begin{abstract}

End-to-end autonomous driving aims to build a fully differentiable system that takes raw sensor data as inputs and directly outputs the planned trajectory or control signals of the ego vehicle. State-of-the-art methods usually follow the `Teacher-Student' paradigm. The Teacher model uses privileged information (ground-truth states of surrounding agents and map elements) to learn the driving strategy. The student model only has access to raw sensor data and conducts behavior cloning on the data collected by the teacher model. By eliminating the noise of the perception part during planning learning, state-of-the-art works could achieve better performance with significantly less data compared to those coupled ones.

However, under the current Teacher-Student paradigm, the student model still needs to learn a planning head from scratch, which could be challenging due to the redundant and noisy nature of raw sensor inputs and the casual confusion issue of behavior cloning.  In this work, we aim to explore the possibility of directly adopting the strong teacher model to conduct planning while letting the student model focus more on the perception part. We find that even equipped with a SOTA perception model, directly letting the student model learn the required inputs of the teacher model leads to poor driving performance, which comes from the large distribution gap between predicted privileged inputs and the ground-truth. 

To this end, we propose \emph{DriveAdapter}, which employs adapters with the feature alignment objective function between the student (perception) and teacher (planning) modules. Additionally, since the pure learning-based teacher model itself is imperfect and occasionally breaks safety rules, we propose a method of action-guided feature learning with a mask for those imperfect teacher features to further inject the priors of hand-crafted rules into the learning process. DriveAdapter achieves SOTA performance on multiple closed-loop simulation-based 
% self-driving 
benchmarks of CARLA. 
% We will make the code and model publicly available.
\end{abstract}

%1. explain
%2. enjoy the advance of the BEV perception 
%

%%%%%%%%% BODY TEXT
\section{Introduction}\label{sec:intro}
In recent years, autonomous driving has become an active research topic due to the enormous progress of deep learning. A traditional pipeline of an autonomous driving system is usually composed of object detection~\cite{li2022bevformer}, motion prediction~\cite{9363585,pmlr-v205-jia23a}, trajectory planning~\cite{Renz2022CORL}, \textit{etc}. To fully unleash the power of deep learning and big data and avoid cumulative errors, the concept of end-to-end autonomous driving is proposed~\cite{pomerleau1988alvinn,muller2005off,hu2023_uniad,chen2023e2esurvey} which aims to build a fully differentiable model directly mapping the raw sensor data into planned trajectories or control signals. 

One difficulty of end-to-end autonomous driving is that the noisy and redundant raw sensor inputs make it hard to directly learn a good policy. 
For example, raw sensor inputs based reinforcement learning (RL) agent MaRLn~\cite{toromanoff2020end} requires 20 million steps (around 20 days) to converge even equipped with their pretraining techniques. 
To this end, in Roach~\cite{zhang2021roach}, they decouple the learning process into two steps: (i) conduct the RL algorithm based on privileged inputs - rasterizing the ground-truth location of surrounding agents and traffic signs into 2D bird's-eye-view (BEV) tensors. The trained RL model is called the \emph{teacher model} as it uses privileged inputs and thus performs well. (ii) conduct behavior cloning with raw sensor inputs on the data collected by the teacher model. This model only with access to raw sensor data is called the \emph{student model} since they are supervised by the teacher model.  By decoupling the perception noise from the driving strategy learning process, Roach could achieve much better performance on more challenging benchmarks in 10 million steps. 
Besides the benefits of efficient RL training, in LBC~\cite{chen2020learning} and PlanT~\cite{Renz2022CORL}, they demonstrate that training a teacher model from rule-based expert and then using the teacher model to provide extra supervision for the student model could bring significant performance gains as well.
Due to the aforementioned advantages of the decoupled planning and perception learning, the teacher-student paradigm has been widely adopted by the state-of-the-art (SOTA) works~\cite{chen2020learning,chen2022lav,wu2023PPGeo,Renz2022CORL,hu2022model}.

However, there are still issues under the existing paradigm. The student model still needs to train a planning head from scratch by behavior cloning, which could result in the causal confusion issue~\cite{wen2020fighting}.
Specifically, the causal confusion issue here refers to the phenomenon that the student model learns the visual clue of the results instead of the cause of the desired actions. For example, the well-known inertia problem~\cite{wen2020fighting} is that the agent sometimes keeps still forever at the intersection. It is because, during behavior cloning, the student model might learn the improper causal correlation that the ego vehicle should copy behaviors of its surrounding vehicles at the intersection. In fact, the behaviors are determined by the traffic light. However, since the traffic light is smaller in images compared to vehicles, the student tends to find the shortcut~\cite{wen2020fighting}. As a result, during evaluation, if there are no vehicles nearby or all vehicles are behind the ego vehicle, it might get stuck.
The causal confusion issues could be solved by techniques such as reweighting the distribution of training data~\cite{ross2011reduction,prakash2020exploring,wen2020fighting} or a causal prior structure/network~\cite{wen2022fighting,chuang2022resolving}.

%Instead of tackling this issue after it happened, 
Inspired by the fact that we already have a strong teacher model trained by RL without any causal confusion issue, in this work, we aim to explore the way to \emph{utilize the teacher model to conduct planning directly instead of training a planning head for the student model from scratch}.
In this way, the learning process of perception and planning is completely decoupled and thus the disadvantage of behavior cloning could be avoided, as demonstrated in Fig.~\ref{fig:teaser}. As a result, we could directly benefit from the driving knowledge inside the teacher model learned by RL.
One intuitive implementation of this idea is to train a student model to generate the required privileged inputs for the frozen teacher model, \textit{e.g.}, a BEV segmentation student model for the Roach teacher model.
However, we find that even equipped with the SOTA perception model: BEVFusion~\cite{liu2022bevfusion} + Mask2former~\cite{Cheng_2022_CVPR}, its final driving performance is still unsatisfying. The issue comes from the large distribution gap between the predicted BEV segmentation and ground-truth. It could be formulated as a domain transfer problem since the teacher model has only seen the ground-truth BEV segmentation during the training process. 

Inspired by the usage of adapters in the natural language processing (NLP)~\cite{houlsby2019parameter} and computer vision~\cite{gao2021clip} field to adopt huge foundation models for downstream tasks, we propose \textbf{DriveAdapter}, which connects the output of the student model (perception) and the input of the teacher model (planning). Specifically, we add a learnable adapter module after each part of the teacher model and apply feature alignment objective functions on each adapter.  In this way, the adapter could learn to transfer the imperfect feature from the student model's domain to the teacher model's domain in a layer-by-layer supervised way.

Additionally, we observe that the pure learning-based teacher model itself is imperfect and it is a common practice to add extra hand-crafted rules during the final decision process~\cite{wu2022trajectoryguided,zhang2022mmfn,hu2022model}. %\footnote{ https://github.com/OpenPerceptionX/TCP}%, if there are some objects in front of the ego vehicle, it would brake immediately regardless of the actual model output.}. 
Thus, even if the student with adapters could losslessly generate the required inputs for the teacher model, it is still upper-bounded by the imperfect performance of the teacher. To this end, we propose to backpropagate an action loss to all adapters and mask all feature alignment loss if the teacher model is overridden by the rule. In this way, we force adapters to directly learn the feature required to generate good actions instead of just mimicking the teacher.
By combining the two proposed techniques, DriveAdapter achieves state-of-the-art performance on two closed-loop evaluation benchmarks of the CARLA simulator. Moreover, we conduct thorough ablation studies and give results of other related attempts such as directly generating intermediate features of the teacher.

In summary, this work has the following contributions:
\begin{itemize}[leftmargin=*,itemsep=0pt,topsep=4pt]
    \item To the best of our knowledge, we are the first to thoroughly explore the paradigm of directly utilizing the teacher head to conduct planning for the end-to-end autonomous driving task. Under such decoupled paradigm, the disadvantages of behavior cloning such as causal confusion could be avoided.
    
     \item The intermediate output form of BEV segmentation between the perception and planning modules has strong interpretability, shedding insights into the typically black-box pipeline of end-to-end autonomous driving. The decoupled perception model could enjoy the recent rapid progress of BEV perception and semantic segmentation.
    
    \item To deal with the imperfect perception issue as well as the imperfect teacher model issue, we propose DriveAdapter along with a masked feature distillation strategy. By combining the two techniques, it could achieve state-of-the-art performance on two public benchmarks.
    \item We give thorough ablation studies and other related attempts to provide more insights and understanding regarding the new decoupled paradigm. 
\end{itemize}

\emph{We believe that the rich driving knowledge within an RL expert model learned by millions of steps of exploration should be utilized more extensively instead of only for behavior cloning. We hope the proposed decoupled paradigm,  our failing attempts, and our working techniques could all provide useful insights for this line of study}.

\section{Related Works}
\subsection{End-to-End Autonomous Driving}
The concept of end-to-end autonomous driving could date back to 1980s~\cite{pomerleau1988alvinn}. In the era of deep learning, early works conduct behavior cloning from a rule-based expert. CIL~\cite{codevilla2018cil} adopts a simple CNN to directly map the front-view image from a camera to control signals. Further, in the extending work CILRS~\cite{codevilla2019exploring}, they add an auxiliary task to predict the ego vehicles' current speed to alleviate the inertia issue. Later, in MaRLn~\cite{toromanoff2020end}, they explore the way to apply reinforcement learning to obtain a driving policy to surpass the rule-based expert. However, their method suffers from the high-dimensional raw sensor inputs for urban driving. In LBC~\cite{chen2020learning}, they propose to first train a teacher model with privileged inputs and then utilize this teacher model to provide supervision signals for all high-level commands, which significantly boosts the performance of the student model and thus the teacher-student paradigm has dominated the field. Roach~\cite{zhang2021roach} trains the teacher model with reinforcement learning, which demonstrates strong robustness and diversity compared to imitation learning-based teacher models and has been adopted by multiple most recent SOTA works~\cite{wu2022trajectoryguided,hu2022model}. In PlanT~\cite{Renz2022CORL}, they propose to adopt Transformer for the teacher model on the states of the environment instead of CNN on the rasterized images, which demonstrates good scalability and interpretability. As for the student models, NEAT \cite{chitta2021neat} transfers representations to BEV space. Transfuser~\cite{Prakash2021CVPR,Chitta2022PAMI} adopts Transformer for camera and LiDAR fusion. LAV~\cite{chen2022lav} adopts PointPainting~\cite{vora2020pointpainting} and proposes to predict all agents' future trajectories to augment the dataset. TCP~\cite{wu2022trajectoryguided} combines the trajectory prediction~\cite{10192373,pmlr-v164-jia22a} with the control signal prediction. Interfuser~\cite{shao2022interfuser} injects safety-enhanced rules during the decision-making of the student models. MMFN~\cite{zhang2022mmfn} adopts VectorNet for map encoding and MILE~\cite{hu2022model} proposes to learn a world model for the student model. Among concurrent works, ThinkTwice~\cite{jia2023thinktwice} proposes a DETR-like scalable decoder paradigm for the student model. CaT~\cite{zhang2023coaching} designs a knowledge distillation framework for the Teacher-Student paradigm. ReasonNet proposes specific modules for student models to better exploit temporal and global information. In \cite{Jaeger2023ICCV}, they propose to formulate the output of the student as classification problems to avoid averaging. 

We could observe that state-of-the-art works all use behavior cloning for student models. In this work, we explore the way to further decouple perception and planning learning by directly adopting the frozen teacher model for planning, while keeping the system end-to-end differentiable.

\subsection{Adapter for Deep Learning Models}
In recent days, huge foundation models~\cite{bommasani2021opportunities} pretrained on an enormous amount of data have demonstrated strong transfer ability on downstream tasks. Since finetuning the whole model is computationally expensive and tends to overfit on downstream tasks with limited data, 
Adapter~\cite{houlsby2019parameter} is first proposed in the NLP field where they fix the parameters of the original model and add extra learnable parameters between blocks of the original model so that it could keep the generalization ability of the original model while having task-specific changes. 
Later, the Adapter is also shown to be effective in the computer vision field~\cite{gao2021clip,zhang2021tip}.

In this work, we adopt the idea of Adapter to keep the knowledge in the teacher model while filling the gap between the predicted privileged inputs and the ground-truth.

\section{Method}

\subsection{Student Model for Perception Learning}
Suppose we have a teacher model with privileged information as inputs. One popular form of encoding the privilege information is the 2D bird's-eye-view (BEV) tensors~\cite{chen2020learning,zhang2021roach} composed of the rasterized position of surrounding agents, lanes, and traffic signs, where the value of each channel could be either 0 or 1 to represent the existence of the corresponding type of object at certain locations (In PlanT~\cite{Renz2022CORL}, they propose to encode the scene into discrete tokes so that Transformer could be adopted). Here, we choose Roach to match with SOTA works~\cite{wu2022trajectoryguided,hu2022model}.

The student model takes raw sensor data as inputs (in this paper, images from four cameras and the point cloud from one LiDAR) and generates the desired input for the teacher model. For Roach, it could be formulated as a semantic segmentation task under BEV space~\cite{pan2020cross}. Here, we adopt the BEVFusion~\cite{liu2022bevfusion} to convert raw sensor data into BEV features. Specifically, we adopt LSS~\cite{philion2020lift} to scatter the image features from cameras into their corresponding BEV grid based on their location and depth~\cite{li2022bevdepth}. For LiDAR, we adopt the commonly used SECOND~\cite{yan2018second} backbone to convert the point cloud into a BEV feature map. By concatenating the BEV feature maps of cameras and LiDAR, we obtain the 2D BEV representation of the scene. To conduct the semantic segmentation task, we adopt the state-of-the-art Mask2former~\cite{Cheng_2022_CVPR} head.

\begin{table}[]
% \small
\centering
\scalebox{0.9}{
\begin{tabular}{ccc}
\toprule
\textbf{Method}                                                          & \textbf{Input}          & \textbf{Driving Score} $\uparrow$ \\ \midrule
Transfuser~\cite{Prakash2021CVPR,Chitta2022PAMI} & Camera + LiDAR & 31.0 \\
LAV~\cite{chen2022lav} & Camera + LiDAR & 46.5 \\
\begin{tabular}[c]{@{}c@{}}Student Model\\ + Frozen Roach\end{tabular} & Camera + LiDAR     &   8.9                     \\ \midrule
Roach~\cite{zhang2021roach}                                                                  & Privileged Info. & 74.2                   \\
Roach + Rule~\cite{wu2022trajectoryguided}                                                            & Privileged Info. & \textbf{87.0}                   \\
\bottomrule
\end{tabular}}
\vspace{-5pt}
\caption{{\textbf{Performance comparison among different teacher and student models on Town05 Long.} 
 Refer to Sec.~\ref{sec:exp} for details of methods and metrics.}
\label{tab:teacher-student}
% \vspace{-5mm}
}
\end{table}

Nevertheless, even equipped with the most advanced perception module as of today, we find that directly feeding the predicted BEV segmentation (mIoU 0.35 on test unseen scenes) to the teacher model does no work, \textit{i.e.}, much worse driving performance compared to SOTA works as shown in Table~\ref{tab:teacher-student}.% (Please refer to Sec.~\ref{sec:exp} for experiment details)

\begin{figure}[!tb]
		\centering
  \includegraphics[width=0.99\linewidth]{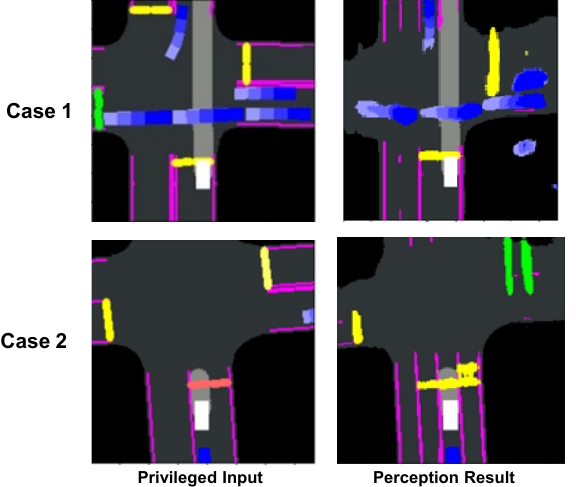}
		\caption{\textbf{Visualization of BEV segmentation results.} The distribution gap between the ground truth and perception results is obvious. Note that the routes and the ego vehicle are directly painted instead of being predicted.\vspace{-5mm}}
\label{fig:vis-bev-seg}
\end{figure}

\begin{figure*}[th!]
    \centering
    \includegraphics[width=\textwidth]{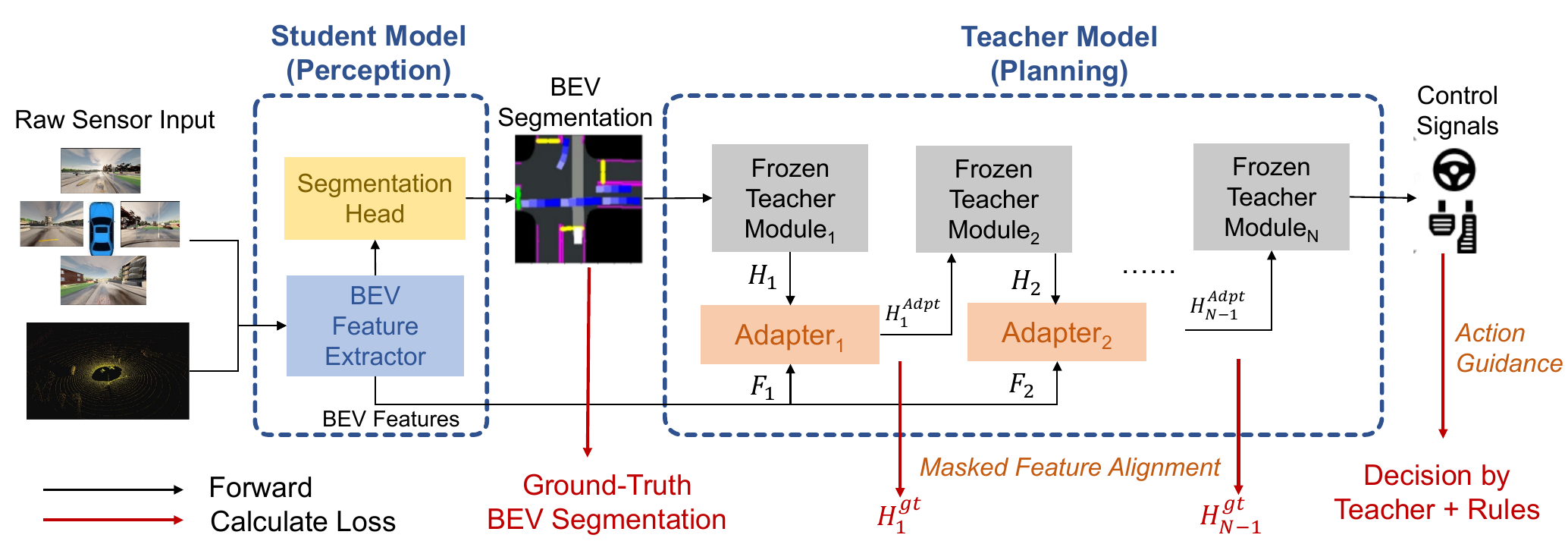}
    \vspace{-20pt}
    \caption{\textbf{Overall architecture of DriveAdapter.} (a) The student model takes raw sensor data as inputs and extracts BEV features for the usage of BEV segmentation and adapter module. (b) The predicted BEV segmentation is fed into the frozen teacher model and the plug-in adapter module. (c)  The adapter module receives supervision from the feature alignment objective with the ground-truth teacher feature. For cases the teacher model is taken over by rules, a mask is applied on the alignment loss and the supervision of all adapter modules is from the backpropagation of action loss.
    %\vspace{-10pt} 
    \label{fig:overall}}
\end{figure*}

There are two factors that impede the performance of the student model + frozen Roach:
\begin{itemize}[leftmargin=*,itemsep=0pt,topsep=4pt]
    \item \emph{Imperfect perception results}. Though we have carefully tuned the BEVFusion + Mask2former on the CARLA scene, the predicted BEV segmentation still has a relatively large gap compared to the ground-truth. As illustrated in Fig.~\ref{fig:vis-bev-seg}, the predicted surrounding agents' location and lanes are incomplete and blurry. Worse still, the states of traffic lights could also be wrong. As a result, the teacher model could not handle such inputs since it has only seen the ground-truth during its training process.
    \item \emph{Imperfect teacher model.} It is a common practice in the community~\cite{zhang2022mmfn,wu2022trajectoryguided,hu2022model} to apply hand-crafted rules on top of the teacher models' outputs for situations such as emergency braking as the learning-based teacher model would make mistakes as well. As shown in Table.~\ref{tab:teacher-student}, after adding rules to the Roach, we observe its performance is boosted by a large margin. Therefore, even if the student model could generate perfect BEV segmentation, its performance is still upper bounded.
\end{itemize}

\subsection{Adapter Module}

% results, it could be viewed as a kind of domain transfer problem where
%as for the issue of imperfect perception issue where
Since the teacher model has only been trained by the BEV segmentation ground-truth, it is sensitive to the noise in the predicted BEV segmentation because of the large distribution gap. Inspired by the usage of the Adapter~\cite{houlsby2019parameter,gao2021clip,zhang2021tip} on foundation models~\cite{bommasani2021opportunities} to adopt them into downstream tasks with lower cost and less overfitting, we propose to add Adapters between the student and teacher model. The overall architecture of DriveAdapter is shown in Fig.~\ref{fig:overall}.  

Formally, denote the predicted BEV segmentation as $\bm{H}_0$ where the subscript $0$ means that they could be the initial input of the teacher model. Suppose the teacher model has $N$ modules in a sequential order\footnote{For more complex teacher models, specific adapter module could be designed accordingly.}, where $\text{Teacher}_i$ denotes its $i^{\text{th}}$ module and $\bm{H}_{i-1}$ and $\bm{H}_i$ denote the original input and output of $\text{Teacher}_i$ respectively: $\bm{H}_i = \text{Teacher}_i(\bm{H}_{i-1})$. Besides, since we want the adapter to have access to the raw sensor inputs so that the model could enjoy the benefits of end-to-end learning, we denote the raw BEV feature from the student model %(feature from BEVFusion, in this work)
as $\bm{F}$ and we use a series of convolutional layers to downsample $\bm{F}$ so that $\bm{F}_i$ could have the same resolution with $\bm{H}_i$.

\begin{figure}[th!]
    \centering
    \includegraphics[width=\linewidth]{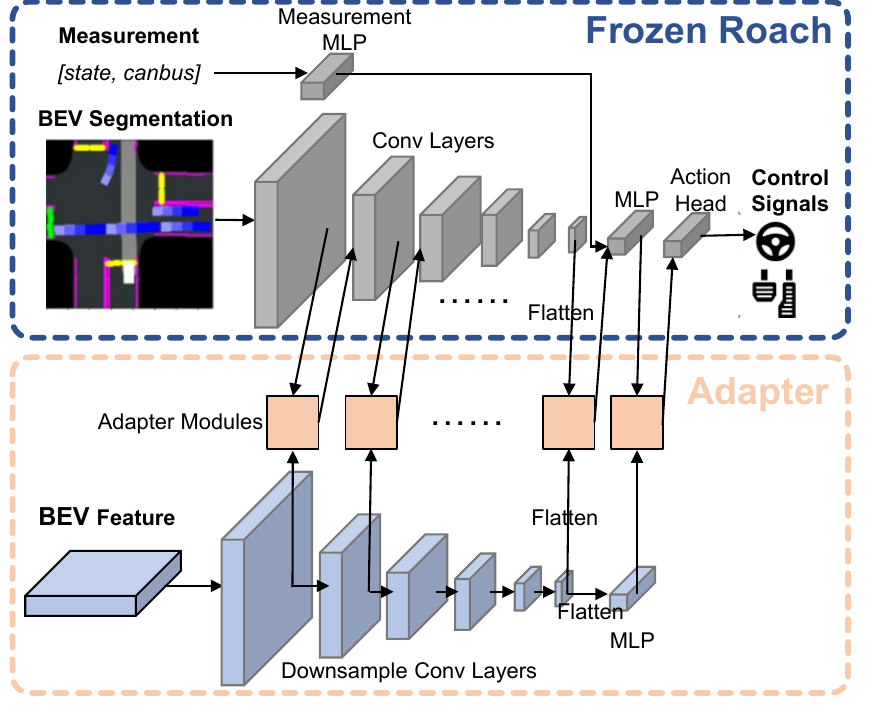}
    \vspace{-15pt}
    \caption{\textbf{Details of Adapter Modules wtih Roach.}}
    \vspace{-10pt} 
    \label{fig:adapter}
\end{figure}

\hspace{-13pt}\textbf{Adapter:} The forward process of the frozen teacher model with adapter modules is:
\begin{align}
   \bm{H}_{i-1}^\text{Adpt} &= \text{Adapter}_{i-1}([\bm{H}_{i-1};\bm{F}_{i-1}]),  \label{equ:adapter1} \\
   \bm{H}_i &= \text{Teacher}_i(\bm{H}_{i-1}^\text{Adpt}), \label{equ:adapter2}
\end{align}
where $\text{Adapter}_{i-1}$ in 
Eq.~\ref{equ:adapter1} denotes the adapter module for $i^{th}$ module of the teacher model which takes the feature from the previous layer $\bm{H}_{i-1}$  and the raw feature $\bm{F}_{i-1}$ as inputs and outputs the adapted feature for the next layer $\bm{H}_{i-1}^\text{Adpt}$ which has exactly the same tensor shape with $\bm{H}_{i-1}$. The  $\text{Adapter}_{i-1}$ is implemented as CNN layers for 2D feature maps and as multi-layer perceptrons (MLPs) for 1D feature maps. Specifically, with Roach as the teacher model, after each layer of Roach’s network (both convolutional layers and linear layers), there is a
corresponding adapter module and we apply feature alignment loss on each output. There are two exceptions: (i) the measurement encoder because it takes the state of the ego vehicle as inputs which is provided directly by the sensor and thus there is no error; (ii) the output linear layer since it generates actions instead of features. Fig.~\ref{fig:adapter} gives the details of the adapter modules.

\vspace{8pt}
\hspace{-13pt}\textbf{Feature Alignment:} To fill the gap between the student's prediction and the ground-truth inputs for the teacher model, we apply a \emph{feature alignment objective function} for each adapter module:
\begin{equation}
    \mathcal{L}_i = \text{Reg}( \bm{H}_{i-1}^\text{Adpt}, \bm{H}_{i-1}^\text{gt}), \label{equ:feature-alignment}
\end{equation}
where $\bm{H}_{i-1}^\text{gt}$ denotes the feature from $(i-1)^{th}$ module of the teacher model (without the adapter module) when the input is the ground-truth BEV segmentation. \text{Reg} denotes the regression loss function and we simply adopt smooth L1 loss here. The intuition behind this design is that we want each adapter module to recover the ground-truth feature required by the teacher model with an additional information source - the raw BEV feature. In this way, the distribution gap between the prediction and the ground-truth feature could be reduced gradually in a layer-by-layer supervised way.

\vspace{3pt}
\hspace{-13pt}\textbf{Mask \& Action Guidance:} As for the imperfect teacher model issue, we inject the priors of the hand-crafted rules into the training process in two ways: (i) \emph{Mask for Feature Alignment}: For the cases where the teacher model is wrong and is taken over by the rules, masks are applied on all feature alignment losses since the original features in the teacher model lead to wrong decisions and thus we did not want the adapter module to recover them. (ii) \emph{Action Guided Feature Learning}:  In order to let the adapter module transform the features of the original teacher model into the features leading to the final decision by rules, we calculate the loss between the model prediction and the actual decision and make it backpropagate all the way through the frozen teacher model and adapter modules. In this way, for those cases, the adapter module is able to turn the final output of the model into the right decision. Experiment results show that the mask and the action guidance could explicitly improve driving performance. 
% Fig.~\ref{fig:overall} gives an overall view of DriveAdapter.

\section{Experiments}\label{sec:exp}
\subsection{Dataset and Benchmark} We use the widely adopted CARLA simulator (version 0.9.10.1) for data collection and closed-loop driving performance evaluation. 

As for the data collection, we collect data with the teacher model \emph{Roach}~\cite{zhang2021roach} + \emph{rules} to match with SOTA works~\cite{wu2022trajectoryguided,hu2022model}. Following the common protocol, we collect data at 2Hz on Town01, Town03, Town04, and Town06 for training and the total number of frames is 189K which is similar to~\cite{Prakash2021CVPR,Chitta2022PAMI,chen2022lav,wu2022trajectoryguided}. At each frame, we collect raw sensor data from 4 cameras and 1 LiDAR and we store the labels including depth, segmentation, control signals, ground-truth features, and the states of the ego vehicle. Additionally, recent works have collected more data and even on more towns for training:~\cite{shao2022interfuser} (3 million frames, 8 towns) and ~\cite{hu2022model} (2.9 million frames but at 25 Hz), which might trigger the unfair comparison issue as discussed in the community\footnote{https://github.com/opendilab/InterFuser/issues/3}\textsuperscript{,}\footnote{ https://github.com/wayveai/mile/issues/4}. To this end, we collect a dataset with 2 million frames on 8 towns and we denote all models trained with extra data with *.

As for the evaluation process, we conduct closed-loop running under two public benchmarks: Town05Long (most widely used) and Longest6 (36 challenging routes selected by \cite{Chitta2022PAMI}). Both benchmarks are composed of tens of routes while each route contains a series of target points to indicate the destination of driving. In addition, there are manually defined challenging events randomly happening during driving to evaluate the driving agent's ability to deal with long-tail problems. For example, jaywalking pedestrians might suddenly appear. At the intersection, an opposite vehicle might illegally run a red traffic light. Thus, the model should have comprehensive knowledge about driving instead of simple lane following. Note that during the evaluation process, the model only has access to the raw sensor data and the use of privileged information is prohibited. For implementation details of the simulation and the model, please refer to supplemental materials.

\subsection{Metrics}
Official metrics of CARLA are used. \textbf{Infraction Score (IS)} measures the number of infractions made along the route, with pedestrians, vehicles, road layouts, red lights, \textit{etc}. %\footnote{Please refer to \url{https://leaderboard.CARLA.org/get_started_v1/} for details.} 
\textbf{Route Completion (RC)} is the percentage of the route completed by the autonomous agent.  \textbf{Driving Score (DS)} is the \textbf{main metric} which is the product of Route Completion and Infraction Score.

\begin{table*}[!ht]
\centering
\scalebox{0.9}{
\begin{tabular}{c|cc|c|ccc}
\toprule
%\hline
\textbf{Method} & \textbf{Teacher} & \textbf{Student}      & Reference          & \textbf{DS}$\uparrow$            & RC$\uparrow$           & IS$\uparrow$           \\ \midrule
CILRS~\cite{codevilla2019exploring} & Rule-Based  &   Behavior Cloning     & CVPR 19         & 7.8    & 10.3  & 0.75 \\
LBC~\cite{chen2020learning} & Imitation Learning  &   Behavior Cloning + DAgger     & CoRL 20         & 12.3   & 31.9  & 0.66 \\
Transfuser~\cite{Prakash2021CVPR,Chitta2022PAMI} & Rule-based & Behavior Cloning & TPAMI 22 & 31.0   & 47.5  & \textbf{0.77} \\
Roach~\cite{zhang2021roach} & Reinforcement Learning & Behavior Cloning + DAgger      & ICCV 21         & 41.6   & 96.4  & 0.43 \\
LAV~\cite{chen2022lav}    & Imitation Learning & Behavior Cloning     & CVPR 22         & 46.5   & 69.8  & 0.73 \\
TCP~\cite{wu2022trajectoryguided}   & Reinforcement Learning & Behavior Cloning      & NeurIPS 22      & 57.2  & 80.4 & 0.73 \\
ThinkTwice~\cite{jia2023thinktwice} &   Reinforcement Learning & Behavior Cloning   &  CVPR 23            &  65.0  & 95.5  & 0.69 \\ 
\textbf{DriveAdapter}    & Reinforcement Learning & Frozen Teacher + Adapter   &  Ours            &  61.7  & 92.3  & 0.69 \\ 
\textbf{DriveAdapter} + TCP    & Reinforcement Learning & Frozen Teacher + Adapter   &  Ours            & \textbf{65.9}  &  94.4  & 0.72  \\ 

\midrule
MILE*\dag~\cite{hu2022model}  & Reinforcement Learning &  Model-Based Imitation Learning     & NeurIPS 22      & 61.1  & \textbf{97.4}  & 0.63 \\
Interfuser*~\cite{shao2022interfuser} & Rule-Based & Behavior Cloning + Rule  & CoRL 22         & 68.3 & 95.0 & -            \\
ThinkTwice*~\cite{jia2023thinktwice} &   Reinforcement Learning & Behavior Cloning   &  CVPR 23            &  70.9  & 95.5  & 0.75 \\ 
\textbf{DriveAdapter} + TCP*   & Reinforcement Learning &     Frozen Teacher + Adapter     & Ours              &  \textbf{71.9} & 97.3 & 0.74 \\ 

\bottomrule%\hline
\end{tabular}}
% \vspace{-3pt}
\caption{\textbf{Performance on Town05 Long benchmark.} $\uparrow$ means the higher the better. * denotes using extra data. $\dag$ denotes no scenarios are used, which is a much easier benchmark. \emph{+TCP} means we adopt its dual output technique by adding an additional trajectory prediction head ~\cite{wu2022trajectoryguided}. 
\vspace{-8pt} 
\label{tab:town05-long}}
\end{table*}

\begin{table}[!h]
\centering
\begin{tabular}{c|ccc}
\toprule
\textbf{Method}      & \textbf{DS}$\uparrow$            & RC$\uparrow$            & IS$\uparrow$            \\ \midrule
WOR~\cite{chen2021learning}         & 23.6          & 52.3          & 0.59          \\
LAV~\cite{chen2022lav}         & 34.2          & 73.5          & 0.53          \\ 
Transfuser~\cite{Prakash2021CVPR,Chitta2022PAMI}  & 56.7          &  \textbf{92.3} & 0.62          \\
PlanT with Perception~\cite{Renz2022CORL} &  57.7 & 88.2 & 0.65 \\
ThinkTwice~\cite{jia2023thinktwice} &  61.3 & 73.0 & 0.81 \\
ThinkTwice*~\cite{jia2023thinktwice} &  66.7 & 77.2 & 0.84 \\
\midrule
\textbf{DriveAdapter}  &  59.4    &    82.0      & 0.68 \\
\textbf{DriveAdapter} + TCP  &  62.0        &      82.3     &  0.70      \\ 
\textbf{DriveAdapter} + TCP* &  \textbf{71.4}  &    88.2      &  \textbf{0.85} \\ \bottomrule
\end{tabular}
\caption{\textbf{Performance on Longest6 benchmark.}\label{tab:longest6}}
\end{table}

\subsection{Comparison with State-of-the-Art Works}

We compare with SOTA works on two widely used public benchmarks: Town05 Long and Longest6 as shown in Table~\ref{tab:town05-long} and Table~\ref{tab:longest6} respectively.  We could observe that DriveAdapter performs the best under the limited data setting and after adopting the dual outputs trick in \cite{wu2022trajectoryguided} to improve safety, it even performs on par with competitors trained on 10$\times$ data. After feeding more data to the model, DriveAdapter sets new records on both benchmarks. After investigation, we find that the major gain of 10$\times$ data comes from better detection of the red light - a \emph{perception} issue. This demonstrates the benefit of having an explainable intermediate representation of BEV segmentation. There are also some common issues happening under both limited and 10$\times$ data settings. We give more thorough investigations about failure cases in supplemental materials.%Sec.~\ref{sec:failure}.

\subsection{Ablation Study}
In this section, we conduct ablation studies to verify the effectiveness of each design of DriveAdapter. All experiments are conducted on Town05 Long benchmark with 189K training data.

\begin{table}[!h]
\centering
\begin{tabular}{c|ccc}
\toprule
\textbf{Method}      & \textbf{DS}$\uparrow$            & RC$\uparrow$            & IS$\uparrow$            \\ \midrule
DriveAdapter         & 61.7          & 92.3          & 0.69          \\ \midrule
w/o Feature Alignment Loss & 45.4 &  69.1 & 0.66  \\
w/o Mask for Feature Alignment & 56.9 & 85.4 & 0.65 \\
w/o Action Loss & 47.1 & 90.5 & 0.52 \\
\bottomrule
\end{tabular}
\vspace{-1mm}
\caption{\textbf{{Ablation on loss terms of the adapter.}}\vspace{-6mm} \label{tab:loss-design}}
\end{table}

\subsubsection{Loss Design}
In DriveAdapter, we have two kinds of loss terms for the adapter modules: the feature alignment loss to deal with the distribution gap between the prediction and ground-truth BEV segmentation, and the action loss to handle the cases when the learning-based teacher model makes mistakes and the decision is overridden by rules.

The ablation study regarding the two loss terms is in Table~\ref{tab:loss-design}. We can observe that:
\begin{itemize}[leftmargin=*,itemsep=0pt,topsep=4pt] 
    \item If we do not conduct feature alignment, the supervision for the adapter module is only the action loss which is very similar to behavior cloning. Thus, we could observe a drop in route completion due to the inertia issue.
    \item If we do not apply the masking strategy for those cases where the learning-based teacher model made mistakes, the supervision signal of action loss and feature alignment loss is conflict. As a result, the overall performance drops.
    \item If we discard the action loss, we could observe a drastic drop in the IS (infraction score) which comes from more collisions due to the relatively aggressive teacher model.
\end{itemize}

In summary, we can find out that both loss terms, as well as the masking strategy, are significant for the learning process. The feature alignment loss allows the adapter to exploit the driving knowledge within the teacher model while the action loss and mask strategy inject information about hand-crafted rules which leads to a more conservative and thus safer driving strategy.

\begin{table}[!h]
\centering
\begin{tabular}{c|ccc}
\toprule
\textbf{Method}      & \textbf{DS}$\uparrow$            & RC$\uparrow$            & IS$\uparrow$            \\ \midrule
DriveAdapter         & 61.7          & 92.3          & 0.69          \\ \midrule
Adapter at Early Stage & 47.2 & 93.9 & 0.47 \\
Adapter at Late Stage & 54.3 & 79.9 & 0.69 \\
w/o BEV Raw Feature & 34.8 & 82.3 & 0.43 \\
Unfrozen Teacher Model & 49.0 & 73.2 & 0.72 \\
\bottomrule
\end{tabular}
\vspace{-1mm}
\caption{\textbf{Ablation on the design of the adapter.}\vspace{-4mm} \label{tab:adapter-design}}
\end{table}

\subsubsection{Adapter Design}
\label{sec:adapter-design}
In DriveAdapter, all modules of the teacher model are frozen and after each module, there is an adapter that takes both the feature from the previous layers and the raw BEV features as inputs. In Table~\ref{tab:adapter-design}, we give ablation studies regarding those designs. We can conclude that:
\begin{itemize}[leftmargin=*,itemsep=0pt,topsep=4pt]

    \item If we only set adapters at the early stage of the teacher model - only on 2D feature maps, the agent became very reckless with a significant drop of IS. It might be due to the cumulative errors at the late stage.

    \item If we only set adapters at the late stage - only on the flattened 1D feature maps, the agent stuck more often (lower RC). We conjecture that the lack of details about the scene, \textit{i.e.}, the information at the early stage, makes it hard to fully utilize the causal inference ability of the teacher model and the adapter might serve more as a behavior cloning module.

    \item  If we do not feed the raw BEV feature to the adapters, the agent performs poorly. It is natural since the information source of the downstream module is only the blurry and incomplete predicted BEV segmentation now, which is not enough to recover features.

    \item As expected, unfreezing the teacher model would lead to worse performance. The reason is that the behavior cloning process on the dataset with limited size (compared to tens of millions of steps exploration during reinforcement learning) would cause catastrophic forgetting~\cite{MCCLOSKEY1989109}, which has also been observed in foundation models~\cite{bommasani2021opportunities} when finetuning on one downstream task. In fact, the model's final performance is very similar to LAV~\cite{chen2022lav}, a SOTA behavior cloning-based model. This experiment demonstrates the importance of freezing the teacher model.
\end{itemize}

\begin{table}[!h]
\centering
\begin{tabular}{c|ccc}
\toprule
\textbf{Target}      & \textbf{DS}$\uparrow$            & RC$\uparrow$            & IS$\uparrow$            \\ \midrule
BEV Segmentation         & 8.9          & 93.2          & 0.09          \\
CNN-2         & 16.0          & 88.9          & 0.12          \\
CNN-4         & 19.0          & 88.2          & 0.23          \\
CNN-6         & 36.9         & 94.9          &   0.38        \\
Latent         & 39.0          & 100.0          & 0.39          \\
Action         & 39.2          & 61.2         &  0.62         \\
\bottomrule
\end{tabular}
\vspace{-1mm}
\caption{\textbf{Experiments about different learning targets for the student model.} The driving performance is evaluated by feeding the prediction of the student model directly into the corresponding layer of the teacher model. \emph{CNN-i} denotes the CNN feature map at the $i^{th}$ layer of Roach~\cite{zhang2021roach}. \emph{Latent} denotes the 1D feature map after the linear layer on the flattened BEV feature. \emph{Action} denotes a behavior cloning student model. \vspace{-4mm} \label{tab:target}}
\end{table}

\subsection{Beyond BEV Segmentation}
In this section, we explore the possibility of directly regressing the middle feature maps of the teacher model instead of predicting the BEV segmentation. In other words, the student model does not generate inputs at layer 0 for the teacher model, for example, we let the student generate the feature map at layer 1 of the teacher model and then we feed the predicted feature map into the rest of the frozen teacher model. In this spirit, we conduct experiments with feature maps at different layers of the teacher model and the results are in Table~\ref{tab:target}.

From the results, we find out that as the learning target of the student model becomes deeper, the driving performance increases. We hypothesize that feeding features directly into deeper layers of the teacher model would encounter fewer cumulative errors. The only exception is the pure behavior cloning agent. Since it only has action supervision and does not utilize the teacher model at all, it encounters severe inertia issue which leads to a low route completion (RC).

Nevertheless, as discussed in Sec.~\ref{sec:adapter-design}, features at the early stage contain more detailed information about the scene and some of them might be important for the teacher model to make decisions and the usage of the adapter could alleviate the aforementioned cumulative errors. Thus, in this work, we stick to the BEV segmentation target. Besides, compared to high-dimensional features, the semantic segmentation is human-readable, which could be helpful for debugging the perception issues (\textit{e.g.}, the fog is so heavy that the model could not detect the traffic light far away).

% \hspace{-13pt}\textbf{Failure Case:}  

\section{Conclusion}
In this work, we propose DriveAdapter which could directly utilize the driving knowledge within a teacher model learned via reinforcement learning, in an end-to-end autonomous driving pipeline. To overcome the imperfect perception and the imperfect teacher model issue, we propose the masked feature alignment and action guidance objective function for adapters. DriveAdapter achieves state-of-the-art performance on two closed-loop autonomous driving evaluation benchmarks. We hope this could establish a new direction of research in end-to-end autonomous driving.

\section*{Acknowledgement}
% We thank the reviewers for their constructive comments and suggestions. 
This work was 
% in part 
supported by National Key R\&D Program of China (2022ZD0160104), NSFC (62206172, 62222607), Shanghai Municipal Science and Technology Major Project (2021SHZDZX0102).

%\smallskip
%\noindent\textbf{Limitation \& Future Work:} Since the performance of the teacher model (currently DS 87) is the upper bound of TeacherAdapter, pushing the learning-based teacher model's performance to perfection could benefit the TeacherAdapter as well. Similar to other end-to-end autonomous driving works, DriveAdapter is only evaluated in simulation. Thus, sim-to-real would be a future direction to examine the algorithms in the real world.

{\small
\bibliographystyle{ieee_fullname}
\bibliography{arxiv}
}

\clearpage
\newpage
\appendix

\begin{figure*}[!ht]
    \centering
    \includegraphics[width=0.85\linewidth]{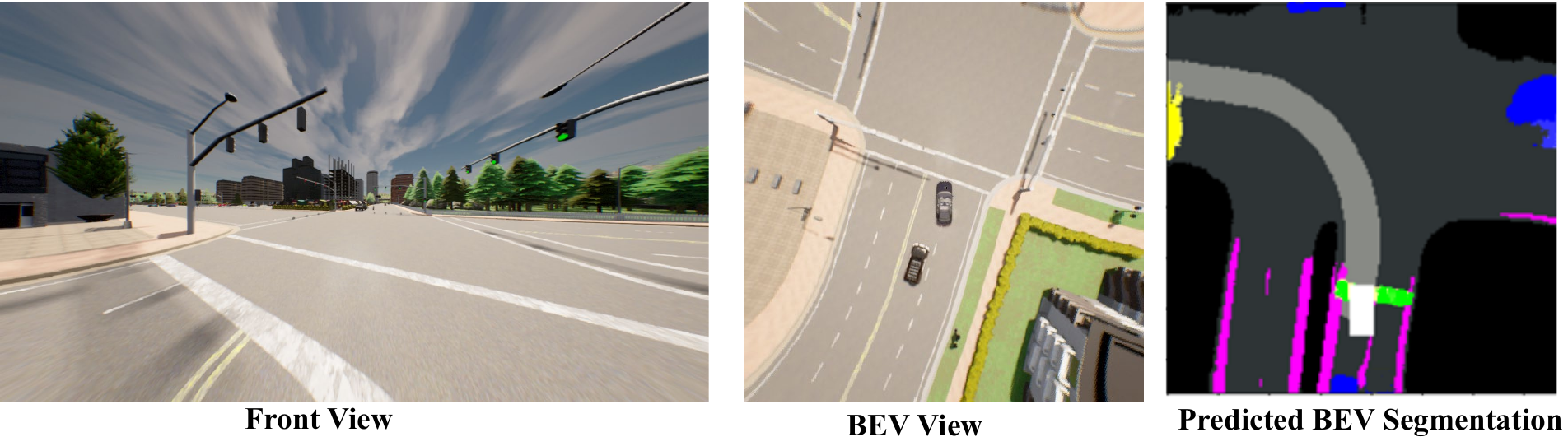}
    \caption{Visualization of the inertia issue where the behavior cloning-based agents often fail. The agent fails to proceed despite the green traffic light, blocking the other car behind.}
\label{fig:vis-success}
\end{figure*}

\section{Case Study of Causal Confusion}  As discussed in the introduction of the paper, behavior cloning based methods suffer from the \emph{inertia issue}~\cite{Chitta2022PAMI}. As shown in Fig.~\ref{fig:vis-success}, at the intersection, if the ego vehicle stops at the front of the line, other vehicles have to wait for its movement. However, since all surrounding vehicles do not move, the ego vehicle would not move since it learns the improper correlation. This is a typical example of \emph{causal confusion}.  

In TeacherAdapter, there is no such situation at all. The route mask in the BEV segmentation is generated by drawing instead of prediction and is fed into frozen the teacher model. Thus, the model naturally has the tendency to move. By comparing with the \emph{Unfrozen Teacher Model} variant in Table 5  in the paper and \emph{Action} variant in Table 6 in the paper , we could find that TeacherAdapter has much higher route completion (RC) score which demonstrates the superiority of utilizing the knowledge within the teacher model over behavior cloning. Also, choosing all learning targets of the student model except \emph{Action} in Table 6 in the paper have high route completion since they all use part of the teacher model.

In summary, by adopting the frozen teacher model for decision-making, the causal confusion issue is avoided and the student model could focus on feature extraction and perception learning. However, there are still some failure cases under this paradigm and we give a thorough investigation of them in the supplemental materials.

\section{Failure Case Analysis}

\begin{figure}[!t]
    \centering
    
    \begin{subfigure}[c]{0.49\textwidth}
    \centering
    \includegraphics[width=0.99\linewidth]{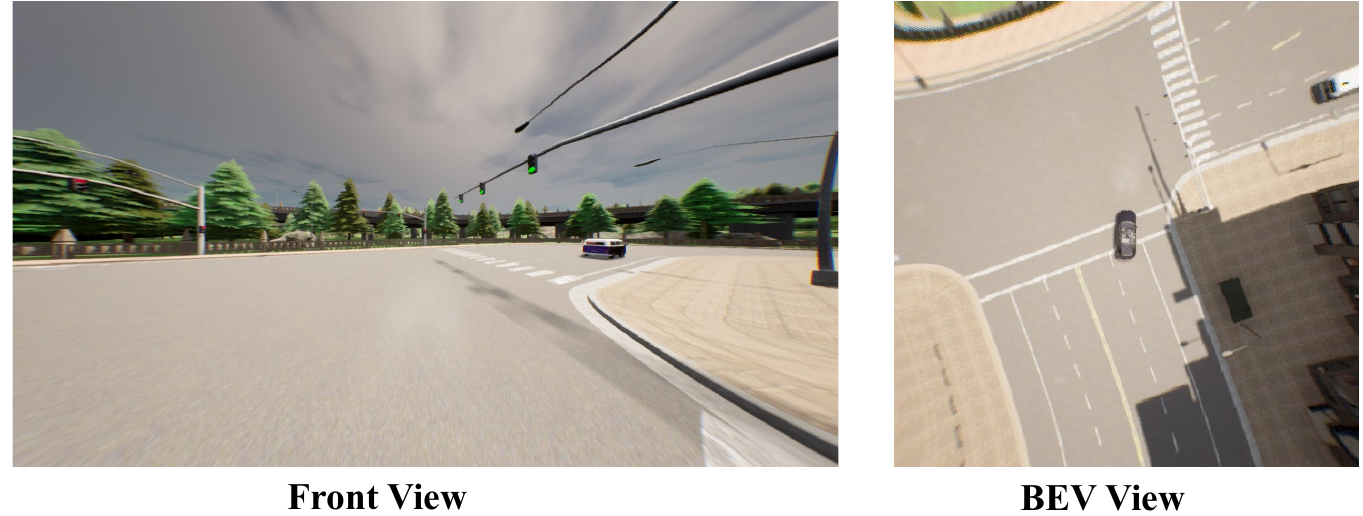}
    \caption{The ego vehicle ran a red light.}
    \end{subfigure}%
    \\
    \begin{subfigure}[c]{0.49\textwidth}
    \centering
    \includegraphics[width=0.99\linewidth]{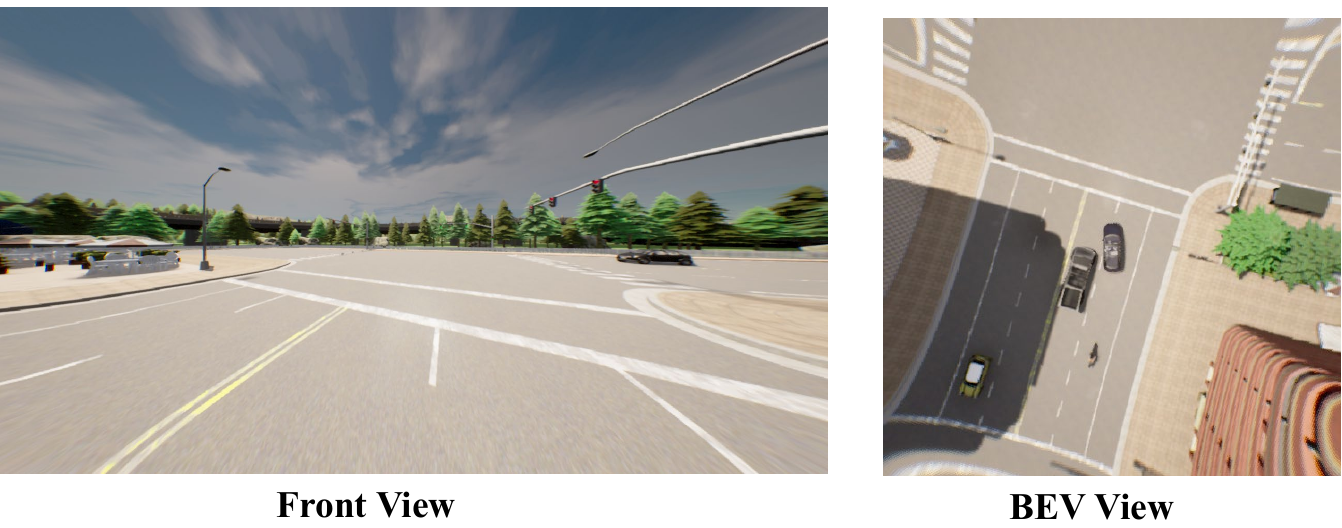}
    \caption{The ego vehicle collided with the vehicle on its back when it started accelerating for a left turn when the traffic light just turned green.}
    \end{subfigure} \\
    \begin{subfigure}[c]{0.49\textwidth}
    \centering
    \includegraphics[width=0.99\linewidth]{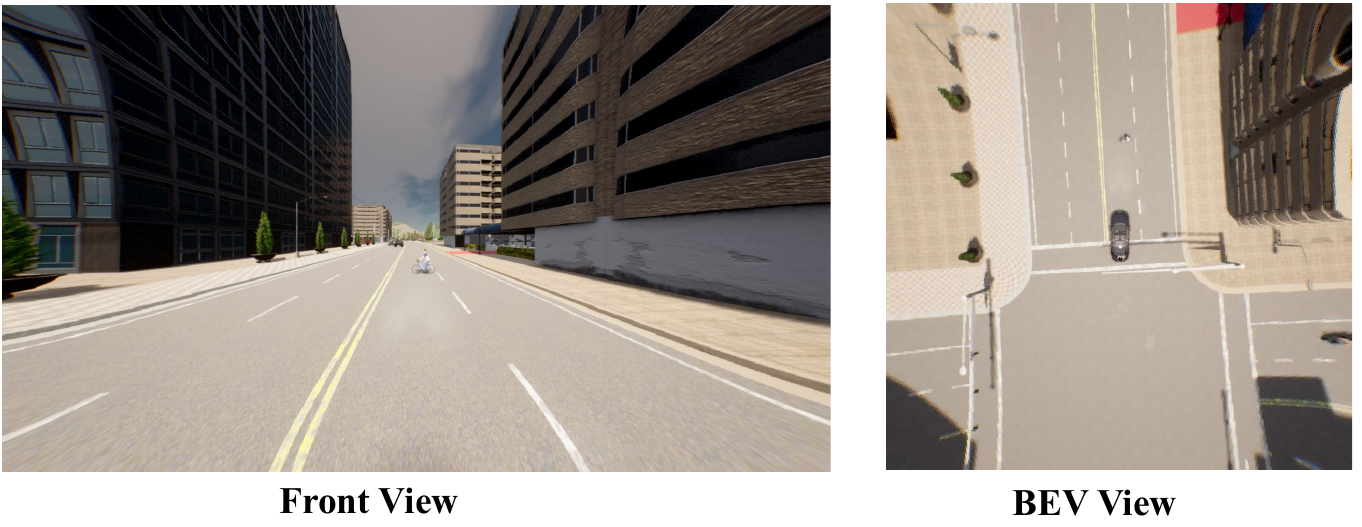}
    \caption{The ego vehicle was stuck since there was a jaywalking pedestrian in front of it.}
    \end{subfigure}

    \caption{\textbf{Visualization of failure cases.}}
    
\label{fig:vis-failure}
\end{figure}

Though DriveAdapter achieves state-of-the-art performance, its performance is still far from perfection (Driving Score 71). For the future development of the work, we summarize three kinds of major failure cases as shown in Fig.~\ref{fig:vis-failure}. We could conclude that:
\begin{itemize} [leftmargin=*,itemsep=0pt,topsep=0pt]
    \item As for the running red light case in Fig.~\ref{fig:vis-failure} (a), the traffic light which controls the ego vehicle's current lane is small in the front camera and the background of the traffic light is green, which makes it extremely difficult to be detected by the model. The traffic light issue has been mentioned frequently in the community~\cite{chen2022lav,hu2022model}. One possible solution is to set a specific camera and a specific neural net to detect the traffic light like in \cite{chen2022lav}. Besides, when we scale up the dataset size (from limited data setting to 10x data setting), the issue is alleviated.
    \item In the collision case at the intersection in Fig.~\ref{fig:vis-failure} (b), the ego vehicle stopped too far from the stop line of the intersection. As a result, when the traffic light turned green, the ego vehicle and its neighbor started accelerating at the same, which results in the collision. Actually, it is a bad habit of the teacher model (Roach) and we find that this kind of collision happens to Roach as well. Thus, a better teacher model could solve the problem.
    \item As for the stuck case in Fig.~\ref{fig:vis-failure} (c), similar to the one in (b), it is due to too much safety distance. This case is a scenario of the benchmark where the simulator would generate a jaywalking pedestrian in front of the ego vehicle. The moving logic of this pedestrian is that it would cross the road when the distance between the ego vehicle and the pedestrian is smaller than a threshold. As a result, for a cautious ego agent, it stop before their distance under the threshold. As a result, both sides can not move and get stuck forever. One solution is to change the logic of the pedestrian to avoid stuck, which might trigger unfair comparisons with other models since it is a public benchmark.  Another solution is to conduct object detection and make the ego agent creep slowly if there is no obstacle around the ego vehicle like in \cite{Chitta2022PAMI}. However, this strategy might cause more violations of rules or collisions sometimes. 
\end{itemize}

In conclusion, though some failure cases are due to the inherent difficulty of perception or the buggy logic of the simulation, the decision-making process still needs improvements for perspectives such as not being too conservative. A better learning-based teacher model could further improve the performance of DriveAdapter.

\section{Discussion about Real World Application of DriveAdapter}
BEV segmentation provides an abstraction of the driving scene which is useful for sim2real setting. Thus, it might be feasible to train a behavior policy with a simulator (for example CARLA) or offline dataset (for example a policy learned on nuPlan or Waymo Motion Prediction data) and finetune it with a perception module and an adapter in the real world.

%%%%%%%%% BODY TEXT
\section{Implementation Details}
Since an end-to-end autonomous driving model is a large system, we provide details of our implementation in terms of data collection, model configuration, training hyper-parameters, and data augmentation. We will make the code and model publicly available.

\subsection{Data Collection}
We use Roach~\cite{zhang2021roach} as the expert with a collision detector for emergency stop similar to~\cite{wu2022trajectoryguided}. 
We set the following sensors: 

We set four cameras with field of view (FOV) $150^\circ$: Front (x=1.5, y=0.0, z=2.5, yaw=$0^\circ$), Left (x=0.0, y=-0.3,  z=2.5,yaw=$-90^\circ$), Right (x=0.0, y=0.3, z=2.5, yaw=$90^\circ$), Back(x=-1.6, y=0.0, z=2.5, yaw=$180^\circ$) where the aforementioned coordinate and angle are all in the ego coordinate system. The output of each camera is a 900x1600 RGB image. Since CARLA~\cite{Dosovitskiy17} simulates the Brown-Conrady distortion~\cite{10.1093/mnras/79.5.384}, we estimate the distortion parameter with the code of~\cite{IBISCape22}. The estimated parameter for the distortion is (0.00888296, -0.00130899,  0.00012061, -0.00338673,  0.00028834) and we use the parameter to 
calibrate images before we feed them into the neural network. We also collected the depth and semantic segmentation label of images. 

We set one Lidar with 64 channels, upper FOV $10^\circ$, lower FOV $-20^\circ$, and frequency 10Hz, following the official protocol. We set it at (x=0.0, y=0.0, z=2.5, yaw=$0^\circ$).

We set an IMU to estimate the yaw angle, acceleration, and angular velocity of the ego vehicle. We set a GPS to estimate current world coordinate of the ego vehicle and a speedometer to estimate current speed of the ego vehicle.

Following the official setting, we also save the target point which might be hundreds meters away as well high-level commands (keep straight, turn left, turn right, etc) provided by the protocol.

To conduct feature alignment, we collect feature maps of Roach at different layers. To conduct action guidance, we store the final action and whether the learning-based model is overridden by the rules.

We convert all raw data into the ego coordinate system.

\subsection{Model Configuration}

Our code is based on  OpenMMLab~\cite{mmcv} with Pytorch~\cite{paszke2019pytorch}, where we use their official implementation of backbones and cooresponding ImageNet pretrained weights if applicable. We use ResNet50~\cite{he2016deep} as the image backbone. We use the PAFPN~\cite{liu2018path} to obtain the multi-scale image features. As for the LSS~\cite{philion2020lift} and depth module, we adopt the code from~\cite{li2022bevdepth}. We use a U-Net~\cite{ronneberger2015u}-like structure for the image semantic segmentation, similar to \cite{chen2022lav,Chitta2022PAMI}. We downsample all images to 450x800 to save GPU memory.  For the Lidar model, we use the SECOND~\cite{yan2018second} implemented by mmdetection3D which consists of HardSimpleVFE, SparseEncoder, SECOND, and SECONDFPN. We use 2 frames as the input. The BEV grid of both modalities is 256x256  and  the scale is (Front=36.8m, Back=-14.4m, Left=-25.6m, Right=25.6m). We conduct the BEV feautres of the two modalities and use a series of CNN to fuse them. Finally,  we only keep the center 192x192 to conduct BEV segmentation which matches with the input of Roach.  To capture the dynamics of surrounding agents so that we could segment past agents' position, in both Lidar
(SECOND) and camera backbones (LSS), we stack 1 extra history frame, which means the input contains two frames [-1, 0]. Specifically, in SECOND, we simply stack all point clouds and add a additional channel to indicate the time-step. In LSS, we transformer the history BEV feature into the current ego coordinate system and concate it with current BEV feature and finally feed them into Conv layers.

As for the BEV segmentation, Roach's privileged inputs have 24 types: road mask, route mask, lane mask, vehicle mask (-3, -2, -1, 0), pedestrian mask (-3, -2, -1, 0), traffic light mask (-3, -2, -1, 0) where (-3, -2, -1, 0) denotes the history timestep in 2Hz. Note that lane and broken lane are represented by 1.0 and 0.5 respectively in the lane mask. Green, Yellow, and Red light are 0.3137, 0.6667, and 1.0 respectively in the traffic light mask. To formulate the task as BEV segmetation, we separate those mixed classes into independent classes and turn the prediction results back when feeding into the teacher model. Additionally, since the route information is given, it is unnecessary to predict the mask and thus we just need to draw the route mask. We adopt the Mask2former~\cite{Cheng_2022_CVPR} for semantic segmentation with the BEV feature as inputs based on their official implementation. We use only one scale (192x192), 3 encoder layers and 6 decoder layers. .We treat each object type at each timestep as
a class and set corresponding queries to conduct segmentation.  In the 189K frames setting, we
only use 4 towns (no overlap with Town05Long and some
overlap with Longest6) to match with Roach, TCP, LAV,
etc. While in the 2M frames settings*, we use all 8 towns (overlap with both Town05Long and Longest6). We observe more accurate BEV segmentation results and less red light infraction in 2M frames settings due to more data and seen towns.

As for the adapter module, for each 2D feature map, the adapter is one Resnet bottleneck with SE~\cite{hu2018squeeze}. For each 1D feature map, the adapter is a two-layer MLP.

As for the \emph{+TCP} setting, we add an MLP with the 1D feature map of the last layer as inputs and use the trajectory generated by the expert as labels.

The total number of parameters
is 135M, the MACs is 1719G, and the inference GPU
memory is around 5G.

\subsection{Hyper-Parameters}
We use AdamW~\cite{loshchilov2018decoupled} optimizer with the learning rate 1e-4, cosine learning rate decay, effective batch size 96, and weight decay 1e-7. We train the model for 60 epochs. For hidden dimensions, we use 256 at most places. For loss weights, we tune them to make sure that each loss is around 1 at the beginning of training. We apply gradient clip based on the L2 norm with the threshold of 35.

\subsection{Data Augmentation}
For data augmentation which we only apply on images, we use the random color transformation similar to~\cite{wu2022trajectoryguided} and random crop before we project image features to the BEV grid.

%%%%%%%%% REFERENCES
% {\small
% \bibliographystyle{ieee_fullname}
% \bibliography{supref}
% }

\end{document}